\documentclass[letterpaper, 10 pt, conference]{ieeeconf}  

\IEEEoverridecommandlockouts                              

\usepackage{amsmath,amsfonts}

\usepackage{array}
\usepackage{textcomp}
\usepackage{url}
\usepackage{verbatim}
\usepackage{graphicx}
\usepackage{cite}
\usepackage{mathrsfs}
\usepackage{bbding}
\usepackage[caption=false,font=normalsize,labelfont=sf,textfont=sf]{subfig}
\usepackage{graphics} 
\usepackage{mathptmx} 
\usepackage{times} 
\usepackage{amssymb}  
\usepackage{multicol}
\usepackage{color,xcolor}
\usepackage{colortbl}
\usepackage{booktabs}
\usepackage{bm}
\usepackage{multirow}
\usepackage{ulem}
\newcommand{\M}[1]{\mathbf{#1}}

\usepackage[mathscr]{euscript}

\title{\LARGE \bf SGNet: Salient Geometric Network for Point Cloud Registration 
}
\author{Qianliang Wu, Yaqing Ding, Lei Luo, Haobo Jiang, Shuo Gu, Chuanwei Zhou, Jin Xie, Jian Yang 
	\thanks{Qianliang Wu, Lei Luo, Haobo Jiang, Shuo Gu, Chuanwei Zhou, Jin Xie, and Jian Yang are with PCA Lab, Key Lab of Intelligent Perception and Systems for High-Dimensional Information of	Ministry of Education, and Jiangsu Key Lab of Image and Video Understanding for Social Security, School of Computer Science and Engineering, Nanjing University of Science and Technology.}
 \thanks{Yaqing Ding is with Visual Recognition Group, Faculty of Electrical Engineering, Czech Technical University in Prague, Prague, Czech Republic.}
	\thanks{E-mail:\tt\small\{wuqianliang,dingyaqing,jiang.hao.bo,shu
 ogu,cwzhou,csjxie,csjyang\}@njust.edu.cn,\tt\small{luolei
 pitt@gmail.com}}
}

\begin{document}

\maketitle
\thispagestyle{empty}
\pagestyle{empty}

\begin{abstract}
Point Cloud Registration (PCR) is a critical and challenging task in computer vision and robotics. One of the primary difficulties in PCR is identifying salient and meaningful points that exhibit consistent semantic and geometric properties across different scans. Previous methods have encountered challenges with ambiguous matching due to the similarity among patch blocks throughout the entire point cloud and the lack of consideration for efficient global geometric consistency. To address these issues, we propose a new framework that includes several novel techniques. Firstly, we introduce a semantic-aware geometric encoder that combines object-level and patch-level semantic information. This encoder significantly improves registration recall by reducing ambiguity in patch-level superpoint matching. Additionally, we incorporate a prior knowledge approach that utilizes an intrinsic shape signature to identify salient points. This enables us to extract the most salient super points and meaningful dense points in the scene. Secondly, we introduce an innovative transformer that encodes High-Order (HO) geometric features. These features are crucial for identifying salient points within initial overlap regions while considering global high-order geometric consistency. We introduce an anchor node selection strategy to optimize this high-order transformer further. By encoding inter-frame triangle or polyhedron consistency features based on these anchor nodes, we can effectively learn high-order geometric features of salient super points. These high-order features are then propagated to dense points and utilized by a Sinkhorn matching module to identify critical correspondences for successful registration. The experiments conducted on the 3DMatch/3DLoMatch and KITTI datasets demonstrate the effectiveness of our method.
\end{abstract}


\begin{figure}\label{motivation}
	\centering
	\subfloat[Ground Truth]{\includegraphics[width=0.4\linewidth,height=2.5cm]{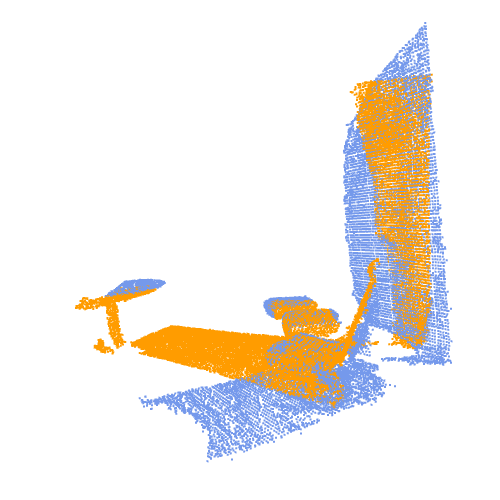}}\quad
	\subfloat[GeoTR~\cite{qin2022geometric}]{\includegraphics[width=0.4\linewidth,height=2.5cm]{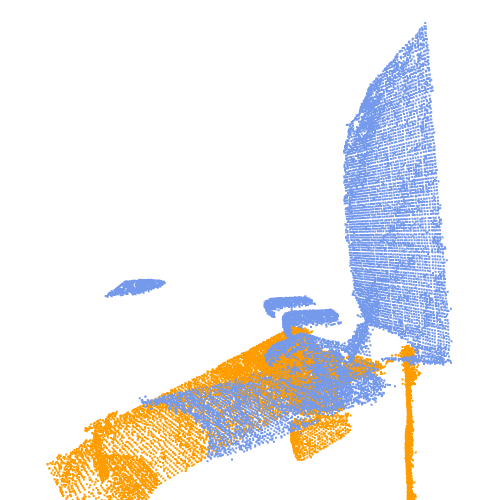}}\quad
	\subfloat[GeoTR~\cite{qin2022geometric}+HO]{\includegraphics[width=0.4\linewidth,height=2.5cm]{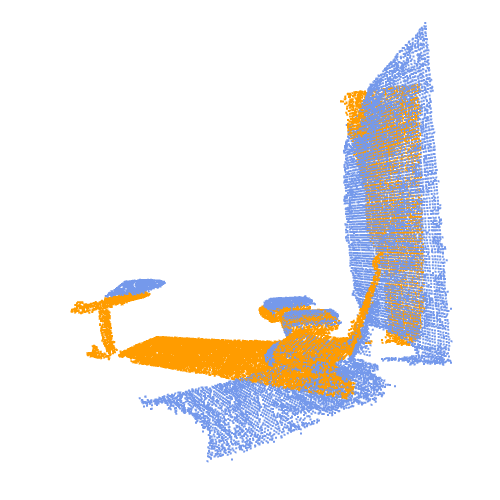}}\quad
	\subfloat[SGNet]{\includegraphics[width=0.4\linewidth,height=2.5cm]{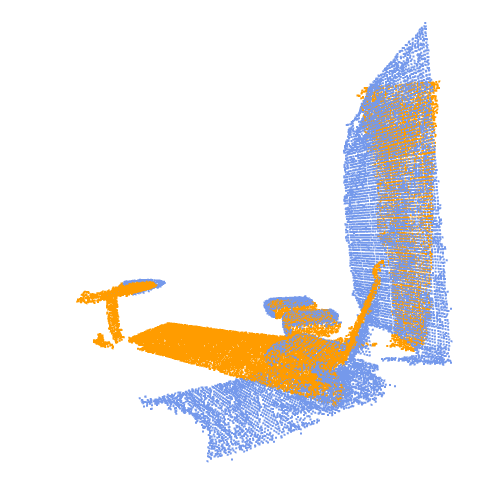}}\quad
	\caption{Our work is inspired by a challenging scenario where successful registration critically depends on the prominent and essential points found on the table, as illustrated in the leftmost region of (a). However, these points are spatially far apart from the abundant points situated on the wall (i.e., points in the rightmost region of (a)). To tackle this challenge, we introduce a new high-order (HO) geometric transformer that leverages high-order geometric features to effectively address such scenarios.}
    \label{hard_case}
\end{figure}

\section{Introduction}
The point cloud registration~\cite{huang2021comprehensive} task is focused on aligning two frames of scanned point clouds. It has numerous applications in computer vision and robotics, including simultaneous localization and mapping (SLAM)~\cite{cadena2016past} and 3D reconstructions~\cite{schonberger2016structure,choi2015robust,zhang2015visual}. Despite progress in 3D point representation learning, detecting salient and repeatable key points across two frames of point clouds for registration remains challenging due to various factors, such as changes in viewpoint, sensor noise, and local ambiguous similarities.

Recent state-of-the-art registration methods mainly adopted the KPFCNN backbone\cite{thomas2019kpconv}. These methods, including~\cite{bai2020d3feat,huang2021predator,yu2021cofinet,qin2022geometric}, initially downsample or divide dense points into coarse key points and extract associated features. They then employ a Transformer~\cite{vaswani2017attention} for information exchange across point clouds. To improve efficiency and accuracy, previous methods such as \cite{yu2021cofinet,qin2022geometric} utilize the point-to-node grouping strategy to divide the dense points into patches. Additionally,~\cite{qin2022geometric} incorporates low-order (i.e., point-wise and pair-wise) and local geometric features at the coarse level to facilitate super point matching, enabling the identification of dense point correspondences within matching patch neighborhoods. 

However, patch-level super points may lead to numerous ambiguous local semantic similarities throughout the point cloud. For instance, down-sampled patches on walls and tables on the ground may exhibit similar regions. This issue may lead to many ambiguous matches. We leverage a pyramid semantic encoder (SE) to encode and exchange the object-level semantic information between point cloud pairs to address this challenge. This encoder can effectively capture high-level object semantics, which can help us to find the salient and meaningful super points for registration.

In challenging real-world scenarios with partial overlapping, salient points are often sparse and widely scattered. Recent methods rely on local~\cite{bai2020d3feat} or low-order~\cite{qin2022geometric} implicit geometric features and may not yield sufficient salient points for matching. Moreover, previous methods that apply self-attention to overlapping and non-overlapping regions can potentially result in the loss of crucial salient points. For instance, consider a complex scene comprising a wall, the ground, several chairs, and a partially obstructed round table (notably, the table is distant from the wall, as depicted in Fig.\ref{hard_case}). In this scenario, there are more inlier corresponding points on the wall or ground compared to the table, leading to the exclusion of table points in~\cite{qin2022geometric,bai2021pointdsc}. Consequently, methods relying solely on local or low-order geometric features may register based only on the wall or ground points, overlooking the importance of table points. It is worth noting that our human eye intuitively detects and selects the points on the wall/ground and table as salient points for alignment. Contrary to the clustered inlier phenomenon, salient points far apart play a critical role in successfully registering challenging scenes. However, the low-order geometric encoder may fail to capture this kind of salient point (e.g., points of the table in (b) of Fig.1). 

To solve this issue, we propose a new high-order geometric feature attention module to select truly salient points. Inspired by hyper-graph matching algorithms~\cite{wang2021neural,lee2011hyper}, we construct two polyhedrons using initial candidate correspondences and compute higher-order similarities by comparing corresponding angles from each polyhedron. Specifically, for each candidate correspondence, we select several distant points within the candidate overlap region to construct two corresponding polyhedrons (for simplicity, triangles are used in this paper). Unlike previous methods that employ naive cross-attention, we introduce a new cross-attention layer to embed the similarities between these two polyhedrons. These high-order features assist in identifying truly salient points within the overlapping regions, thereby contributing to robust registration results in challenging scenes. Furthermore, we propagate these high-order geometric features to the dense points. 

Moreover, it is crucial to have well-initialized, high-quality candidate overlap regions with locally salient points to support downstream modules. Therefore, we employ super points augmentation by leveraging intrinsic shape signatures~\cite{zhong2009intrinsic} to select the most meaningful points throughout the scene as super points. Drawing inspiration from key-point detection methods~\cite{zhong2009intrinsic,bai2020d3feat}, we compute a point saliency score and multiply it with a super points matching score matrix. This process allows us to obtain a candidate overlap region with more salient correspondences. Our contributions are summarized as follows:

\begin{itemize}
 
    \item We introduce a new transformer capable of effectively learning higher-order salient features and distributing them to the dense points, thereby aiding in the learning of dense point features. We also propose an anchor points selection strategy for this new transformer.    
    \item We advocate a novel semantic encoder designed to embed high-level (specifically, object-level) semantic information proficiently. This embedded information is subsequently integrated into a geometric encoder to develop semantic-enriched geometric features for superpoints (at the patch level).

    \item We conduct the experiments on 3DMatch and KITTI datasets and achieve competitive performance.
    
\end{itemize}

\section{Related work}
Recently, many feature-matching methods greatly improved point cloud registration on the recall metric. Most construct their framework on the famous KPConv~\cite{thomas2019kpconv} backbone. Predator\cite{huang2021predator} exploits a GNN network to embed the super point features and utilize a transformer to exchange information between source and target. After that, it propagates the inter-frame exchanged features up to dense points for point-matching. Based on Predator\cite{huang2021predator}, CoFiNet\cite{yu2021cofinet} introduces a point-to-node strategy to group the dense points into downsampled super points' neighborhoods. However, both Predator\cite{huang2021predator} and CoFiNet\cite{yu2021cofinet} suffer from a problem in that they conduct self-attention in the intra-frame on a global scale on both the overlapping and non-overlapping regions. As only overlap regions contain helpful information for successful registration, the transformed super point features, which include non-overlap region information, used by~\cite{huang2021predator} and~\cite{yu2021cofinet}, may lead to ambiguous point matching. To overcome this problem, GeoTR\cite{qin2022geometric} only utilizes a geometric transformer to conduct the patch matching on super points and make dense points matching based on the KPConv features. RoITr\cite{yu2023rotation} designs a new transformer-based feature backbone that utilizes the local PPF\cite{deng2018ppf} features to enhance the rotation invariance.

Furthermore, numerous outlier rejection methods have made significant advancements. For instance, PointDSC\cite{bai2021pointdsc} leverages the max clique algorithm within the neighborhood of keypoint correspondences to identify more dependable inlier correspondences. SC2-PCR\cite{chen2022sc2} exploit second-order consistency graphs that provide more robust inlier correspondences and prove the reliability probabilistically. Based on this second-order consistency graph, MAC\cite{zhang20233d} employs a modified version of the max clique algorithm to further refine the selection of reliable inlier correspondences. Additionally, PEAL\cite{yu2023peal} utilizes a pre-trained GeoTR\cite{qin2022geometric} to generate the overlap region and incorporates simple attention mechanisms and iterative updates to choose more reliable inlier correspondences.

\begin{figure*}[htbp]
      \centering
      \includegraphics[width=\textwidth]{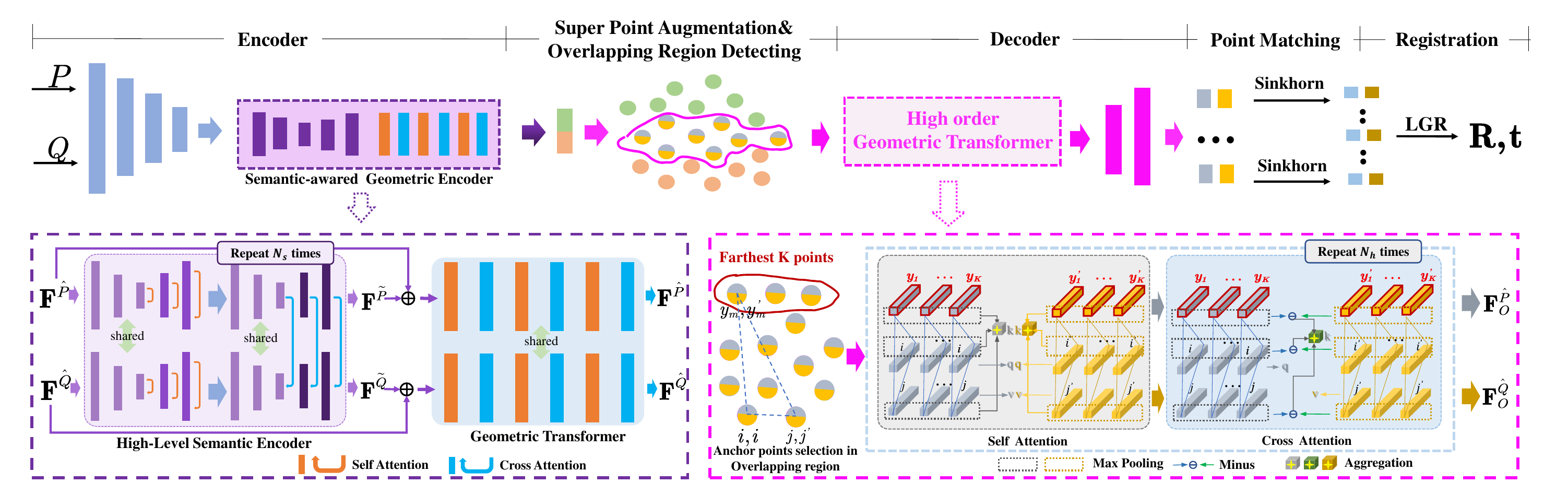}
      \setlength{\abovecaptionskip}{-0.6cm} 
      \caption{Our proposed framework comprises several components. First, the point clouds ${P}$ and ${Q}$ are fed to the down-sampling layers to obtain the super points ($\hat{{P}}$,$\hat{{Q}}$) and their features ($\M F^{\hat{{P}}}$,$\M F^{\hat{{Q}}}$). We then use the semantic-aware geometric encoder to detect overlap regions. Next, our newly designed high-order geometric transformer is applied to encode the high-order inter-frame geometric salient feature ($\M  F^{\hat{{P}}}_o$,$\M  F^{\hat{{Q}}}_o$). We combine the ($\M  F^{\hat{{P}}}$,$\M  F^{\hat{{Q}}}$) and ($\M  F^{\hat{{P}}}_o$,$\M  F^{\hat{{Q}}}_o$) and propagate this combined features up to the dense level points in the overlap region. Finally, the local-to-global registration step computes the 6D rigid transformation ${\M R,\M t}$. Zoom in for details.}
      \label{framework}
\end{figure*}
\section{The proposed Approach}\label{approach}
\subsection{Problem Statement}
Given two frames of point cloud ${P} = \{  {\M p}_i \in \mathbb{R}^3|i=1,...,N\}$ and ${Q} = \{{\M q}_i \in \mathbb{R}^3|i=1,...,M\}$, the point cloud registration aim to find the transformation $\M T \in $ SE(3) (i.e., rotation matrices $\M R \in$ SO(3) and translation vectors $\M t \in {\mathbb{R}^3}$) that aligns ${P}$ to ${Q}$.

\subsection{Method Overview}
Our framework adopts the KPConv\cite{thomas2019kpconv} as the feature extraction backbone, as shown in Fig.\ref{framework}. Firstly, the point clouds $P,Q$ are down-sampled to super points $\hat{{P}}$ and $\hat{{Q}}$, and obtain the associated features ${\M F^{\hat{{P}}}} \in {{\mathbb{R}}^{|{\hat{P}} \times {d_t}|}}$ and ${\M F^{\hat{{Q}}}} \in {{\mathbb{R}}^{|{\hat{Q}} \times {d_t}|}}$. Next, these super points and features are inputted into the high-level semantic encoder to extract object-level semantic-aware patch features $\M F^{\Tilde{{P}}}$ and $\M 
 F^{\Tilde{{Q}}}$. These features are then concatenated with $\M F^{\hat{{P}}}$ and $\M F^{\hat{{Q}}}$ and passed through a low-order and local geometric transformer~\cite{qin2022geometric}, resulting in newly updated super point features $\M F^{\hat{{P}}}$ and $\M F^{\hat{{Q}}}$. Super point matching scores are computed by the inner product of the super point features across two frames. The super points are augmented by utilizing the intrinsic shape signature~\cite{zhong2009intrinsic}. Additionally, salient scores are calculated for each super point and multiplied with the matching scores. The initial candidate overlap region, which represents the correspondences between super point patches, is obtained from the ranked matching score matrix. This process takes place in the Super Point Augmentation $\&$ Overlapping Region Detecting area, located in the middle and at the top of the network architecture. Then we propose a newly designed high-order transformer to learn the high-order inter-frame consistent salient features ($\M F^{\hat{{P}}}_o$,$\M F^{\hat{{Q}}}_o$) and propagate them to dense points. Dense points then utilize these new enhanced features to match patch correspondence neighborhoods to give dense correspondences. Finally, we use a local-to-global estimator to compute the most reliable estimation.

\subsection{High-Level Semantic Encoder}
Although the down-sampled super point contains patch-level semantic information, the local similarities of patches across the whole point cloud may lead to many ambiguous mismatches. To solve this problem, we continue to downsample twice super points $\hat{{P}}$ and $\hat{{Q}}$ to "object level" mega points $\overrightarrow{{P}}$ and $\overrightarrow{{Q}}$ (with associated features $\M F_{\overrightarrow{{P}}}$ and $\M F_{\overrightarrow{{Q}}}$) with larger receptive fields than super points. The magnitudes of dense points (${P}$ and ${Q}$), super points ($\hat{{P}}$ and $\hat{{Q}}$), and mega points ($\overrightarrow{{P}}$ and $\overrightarrow{{Q}}$) are typically in the tens of thousands, hundreds, and tens, respectively. Each mega point resides within the neighborhood of a super point, and each super point resides within the neighborhood of a dense point.

\textbf{Semantic Pooling layers.}
We perform voxel grid subsampling, increasing the cell size at each pooling layer and other related parameters. The subsampled points and associated features in each layer are denoted as $\overrightarrow{{P}}^l|\M F^l_{\overrightarrow{{P}}}$ and $\overrightarrow{{Q}}^l|\M F^l_{\overrightarrow{{Q}}}$ ($l$ = 0,1,2). The points densities of $\overrightarrow{{P}}^2|\overrightarrow{{Q}}^2$ and $\hat{{P}}|\hat{{Q}}$ are same. For efficiency, we utilize a max-pooling to give the features of each new pooled location:
\begin{eqnarray}
f_{\overrightarrow{p_i}^{l}} = {\textit{MAX}}_{\overrightarrow{p_j}\in N_{\overrightarrow{p_i}}} h_{\theta} \left (\textit{CAT}\left[f_{\overrightarrow{p_j}^{l+1}}, f_{\overrightarrow{p_j}^{l+1}} - f_{\overrightarrow{p_i}^{l}}\right]\right)
\end{eqnarray}
where $N_{\overrightarrow{p_i}^l}$ is the radius neighborhood of point $\overrightarrow{p_i}^l$ and ${\overrightarrow{p_j}}^{l+1}$ is a supporting point in this neighborhood. $h_{\theta}$ denotes a linear layer followed by instance normalization and ReLU, \textit{MAX}(·) denotes element-/channel-wise max-pooling, and \textit{CAT}[·, ·] means concatenation. $\M F^l_{\overrightarrow{{Q}}}$ is computed in the same way. Through these aggregation layers, the coarser level points have higher-level semantic information.

\textbf{Upsampling Layers.}
This section describes obtaining object-level semantic information for super points from mega points. Firstly, we update the point feature in layer-$l+1$:
\begin{eqnarray}
f_{\overrightarrow{p_j}^{l+1}} = {h_{\hat{\theta}}} \left (cat\left[f_{\overrightarrow{p_j}^{l+1}}, f_{\overrightarrow{p_i}^{l}}\right]\right), {\overrightarrow{p_j}^{l+1}} \in N_{\overrightarrow{p_i}^l},
\end{eqnarray}
where $h_{\hat\theta}$ is similar with $h_{\theta}$.
Secondly, we utilize a four heads self-attention ($SA(\cdot,\cdot)$) layer on $\M F^l_{\overrightarrow{{P}}}$ and $\M F^l_{\overrightarrow{{Q}}}$:
\begin{eqnarray}
    \M F^l_{\overrightarrow{{P}}} = SA\left(\M F^l_{\overrightarrow{{P}}},\M F^l_{\overrightarrow{{P}}}\right),\ \M F^l_{\overrightarrow{{Q}}} = SA\left(\M F^l_{\overrightarrow{{Q}}},\M F^l_{\overrightarrow{{Q}}}\right),
\end{eqnarray}
We also employ a four heads cross-attention ($CA(\cdot,\cdot)$) layer between $\M F^l_{\overrightarrow{{P}}}$ and $\M F^l_{\overrightarrow{{Q}}}$, respectively:
\begin{eqnarray}
    \M F^l_{\overrightarrow{{P}}} = CA\left(\M F^l_{\overrightarrow{{P}}},\M F^l_{\overrightarrow{{Q}}}\right),\ 
    \M F^l_{\overrightarrow{{Q}}} = CA\left(\M F^l_{\overrightarrow{{Q}}},\M F^l_{\overrightarrow{{P}}}\right).
\end{eqnarray}
In the self-attention $SA(\cdot,\cdot)$, the q, k, and v are computed as:
\begin{eqnarray}
    q_i = \M W_qf_{p_i},&k_j = \M W_kf_{p_j},&v_j = \M W_vf_{p_j}
\end{eqnarray}
where $\M W_q,\M W_k,\M W_v \in \mathbb{R}^{d_t{\times}d_t}$ and $\alpha_{ij} = softmax(q_ik_j^T/\sqrt{d_t})$. The cross-attention $CA(\cdot,\cdot)$ is computed in the same way that $q$ and $k,v$ are computed by source and target point clouds, respectively. 

Between self-attention and cross-attention, we utilize semantic pooling layers to downsample the patch-level point features from the output of self-attention, resulting in mega points. 
More details of self/cross attentions can refer to~\cite{vaswani2017attention}. The output of the semantic encoder is denoted as $\M F_{\tilde{{P}}}$ and $\M F_{\tilde{{Q}}}$.

While mega points are designed to capture object-level semantic information, their large receptive fields often result in a loss of geometric details. To mitigate this issue, we propose a solution by concatenating the object-level semantic-aware super point features ($\M F^{\tilde{P}}$ and $\M F^{\tilde{Q}}$, decoded from mega point features) with patch-level semantic information from the super point features ($\M F^{\hat{P}}$ and $\M F^{\hat{Q}}$). These combined features are then fed into the subsequent geometric transformer as inputs. This fusion process allows us to generate a semantic-aware geometric embedding tailored explicitly for the super point salient features.

\subsection{Super Points Augmentation}\label{ol_detection}
In this section, we describe how we generate the candidate overlapping region. 

\textbf{Salient Superpoint Augmentation.}
To tackle the problem of meaningless super points in the last downsampling layer, which lacks object-level semantic information, we can choose more interesting key point subsets from super points by utilizing keypoint detector algorithms\cite{li2019usip,harris1988combined,lowe2004distinctive,zhong2009intrinsic}. In this paper, we utilize ISS\cite{zhong2009intrinsic} with a spherical neighborhood of radius $r_{salient}$ to choose key points. We apply two threshold values $\lambda_{10}$ and $\lambda_{21}$, and utilize $e[1]/e[0]<\lambda_{10}$ and $e[2]/e[1]<\lambda_{21}$ to select the salient points as new super points. During the upsampling process in the decoder, each superpoint's neighborhood is updated. This step automatically identifies and selects the dense points that describe the object semantics and are the most salient or semantically meaningful within the scene.

\textbf{Salient Super Point Matching.}
We exploit the eigenvalues of the super points' scatter matrix~\cite{zhong2009intrinsic} to compute the salient matching score matrix $\M {SS}\in \mathbb{R}^{|\hat{P}|\times|\hat{Q}|}$. The final 
 super point matching matrix $\M S\in \mathbb{R}^{|\hat{P}|\times|\hat{Q}|}$ is obtained by multiplying these two matrices:
\begin{equation}
\begin{split}
    &\gamma(i)=\left(1-\frac{e_i[1]}{e_i[0]}\right)\left(1-\frac{e_i[2]}{e_i[1]}\right),\\
    &\gamma(j)=\left(1-\frac{e_j[1]}{e_j[0]}\right)\left(1-\frac{e_j[2]}{e_j[1]}\right),\\
    &\M {SS}(i,j) = \gamma(i)*\gamma(j),\\
    &\M S(i,j) = \M {MS}(i,j)*\M {SS}(i,j),
\end{split}
\end{equation}
where $\M {MS}$ is computed by the inner product of the superpoint features across two frames.

After that, we select top $N_c$ super points correspondences in $S$:
\begin{eqnarray}
    \hat{\M C} = \left\{(\hat{{\M p}}_{x_i},\hat{{\M q}}_{y_j})|(x_i,y_j)\in topK_{x,y}\M S(x,y)\right\}
\end{eqnarray}
This initial candidate matching matrix provides sufficient reliable salient correspondences through the augmentation above and salient matching scores.

\begin{figure}[t]
    \centerline{\includegraphics[width=2.0in,height=3.5cm]{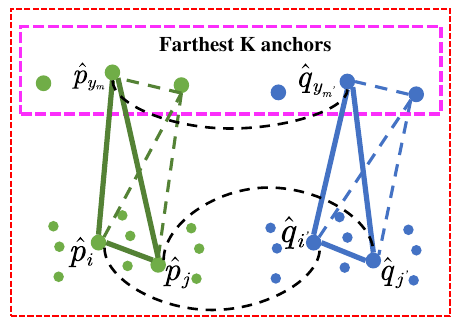}}
    \setlength{\abovecaptionskip}{-0.1cm} 
	\caption{\textbf{Anchor points selection.} The candidate overlap region consists of green and blue points corresponding to each other (indicated by the black dashed curve representing initial candidate correspondences). From this set of points, a subset of anchor points (highlighted within the pinkish-red dashed box) is selected based on their maximum distance from ${\hat{p}}_i$ and ${\hat{q}}_{i^{'}}$.
    In an example scenario where $K=3$, two anchor points generate up to four triangles for ${\hat{p}}_i$ and ${\hat{p}}_j$. Consequently, the angles in these four triangles can be pooled together to aggregate the embeddings for ${\hat{p}}_i$ and ${\hat{q}}_{i^{'}}$.}
    \label{anchor_selection}
  
\end{figure}

\subsection{High Order Geometric Feature}
Despite the super point matching score matrix that can offer candidate matching patches, the dense point matching lacks globally consistent geometric features~\cite{huang2021predator,yu2021cofinet,qin2022geometric}. On the other hand, conducting self-attention on both overlap and non-overlap regions~\cite{qin2022geometric} may introduce ambiguous information for the super point, leading to ambiguous point matching or non-salient correspondence selection. To address this issue, we introduce a new high-order geometric encoder on the candidate overlap region (i.e., superpoint correspondences) to model a robust salient geometric feature and propagate it to dense points by upsampling. The super point learns this high-order salient geometric feature via backpropagation from optimization of the dense point matching loss.

\subsubsection{Anchor Points Selection}
The viewpoint changes and sensor jittering can result in point cloud pairs with partially overlapping regions. After obtaining the candidate super point overlap region, selecting appropriate salient points becomes crucial for successful registration. In real-world scenarios where data is particularly challenging, we observe that key salient points may be widely dispersed from one another, which explains why the clustered inlier correspondences\cite{huang2021predator} contribute little to the registration success. To address this challenge, we deploy the following anchor point selection strategy:
 
\textbf{Farthest $K$ anchor points.}
 As depicted in Fig.\ref{anchor_selection}, for each point (i.e., $\hat{p}_i$), we select the farthest $K$ anchor points in the candidate overlap region. When computing the feature of $\hat{p}_i$ with attention weighting on other points (e.g., $\hat{p}_j$), we assume each point $\hat{p}_i$ has an associate point $\hat{p}_j$. Points $\hat{p}_i$, $\hat{p}_j$, and anchor point $\hat{p}_k$ construct a triangle.

The selected $K$ anchor points will give $\binom{K}{M}$ polyhedrons for $(M+2)-$ order polyhedron similarity. This paper explores the simplest case (i.e., $M=1$) for efficiency, where the $K$ anchor points will give $K$ triangles for each point (e.g., $\hat{p}_i$ and its associated point $\hat{p}_j$). We leave the $M>1$ case in future work.

\subsubsection{High-order Saliency Transformer}
After obtaining the $K$ triangles, we conduct a new transformer to learn the high-order geometric salient features.

\textbf{Triangle spatial consistency encoding.}
For each anchor triangle (e.g., $\Delta(i,j,y_m),m\in \{1, 2,...,K\}$, see bottom right in Fig.\ref{framework}) of the super point pair ($\hat{p}_i$,$\hat{p}_j$), there are three angles $\angle_{ijy_m},\angle_{y_mij}, and \angle_{jy_mi}$. Since we have $K$ anchor points, we design to conduct average pooling on three angles across the $K$ generated anchor triangles:
\begin{eqnarray}
\begin{aligned}
    &{\M g}_{y_m,i,j} = f_{sin}(\angle_{y_mij}),\quad {\M g}_{y_m,j,i} = f_{sin}(\angle_{y_mji}), \\
    &{\M g}_{i,y_m,j} = f_{sin}(\angle_{iy_mj}),\quad {{\M g}_{yij}^{\hat{{P}}}}=max_{y_m} \left \{{\M g}_{y_m,i,j}^{\hat{{P}}}{\M W}^A_1 \right \},\\
    &{{\M g}_{yji}^{\hat{{P}}}}=max_{y_m} \left \{{\M g}_{y_m,j,i}^{\hat{{P}}}{\M W}^A_2 \right \},\quad {{\M g}_{iyj}^{\hat{{P}}}}=max_{y_m} \left \{{\M g}_{i,y_m,j}^{\hat{{P}}}{\M W}^A_3 \right \},
\end{aligned}
\end{eqnarray}
where ${\M W}^A_1, {\M W}^A_2$, and ${\M W}^A_3 \in \mathbb{R}^{d_t \times d_t}$ are the respective weights for encoding of three angles and $m=1...K$. After obtaining three angle embeddings for each point, we utilize them to construct the global self-attention and cross-attention layers. 

\textbf{Self-Attention.}
To encode the intra-frame global scale self-attention in the overlap region, we aggregate the three-angle embeddings:
\begin{eqnarray}
    \begin{aligned}
    &{\bar{\M g}_{ij}}^{self} = Linear\left(sum[{{\M g}_{yij}}, {{\M g}_{yji}}, {{\M g}_{iyj}}]\right),\\
    &e_{ij}^{self} = \frac{{({\M x}_i} \M W^Q)(\M x_j \M W^K + {\bar{\M g}_{ij}^{self}}{\M W^G})^\top}{\sqrt{d_t}},\\
     &{\M z}_i = \Sigma^{|\hat{{P}}|}_{j=1} a_{ij}^{self}({\M x}_j{\M W^V}),
    \end{aligned}
\end{eqnarray}
where ${\M W}^Q, {\M W}^K, {\M W}^V$, and ${\M W}^G \in \mathbb{R}^{d_t \times d_t}$ are the respective projection matrix for key, query, value, and similarity triangle encoding.

\textbf{Saliency Cross-Attention.}
To learn the high-order inter-frame geometric salient features, we design a novel cross-attention layer to encode the high-order saliency feature for points in the candidate overlap region. For each point, $\hat{p}_i$ from the source and $\hat{q}_j$ from the target, we first subtract the corresponding angle embedding between frames as distance in feature space. Then, we aggregate these three feature distances as an inter-frame similarity score (inspired by Eq.(14) in ~\cite{lee2011hyper}):

\begin{eqnarray}
\begin{aligned}
    &{{\bar{\M g}}_{yij}} = ({{\M g}_{yij}^{\hat{{P}}}} - {{\M g}_{yij}^{\hat{{Q}}}})/\sigma_h,\\
    &{\bar{\M g}_{yji}} = ({{\M g}_{yji}^{\hat{{P}}}} - {{\M g}_{yji}^{\hat{{Q}}}})/\sigma_h,\\
    &{\bar{\M g}_{iyj}} = ({{\M g}_{iyj}^{\hat{{P}}}} - {{\M g}_{iyj}^{\hat{{Q}}}})/\sigma_h,\\    
    &{\bar{\M g}_{ij}}^{cross} = Linear\left(concat[{\bar{\M g}_{yij}}, {\bar{\M g}_{yji}}, {\bar{\M g}_{iyj}}]\right),\\
    &e_{ij}^{cross} = \frac{(\M x_i^{\hat{{P}}} \M W^Q)(\M x_j^{\hat{{Q}}} \M W^K + {\bar{\M g}_{ij}^{cross}}{\M W^H})^\top}{\sqrt{d_t}},\\
    &{\M z}_i = \Sigma^{|\hat{ {{Q}}}|}_{j=1}a_{ij}^{cross}{\M x}_j^{\hat{{Q}}}{\M W}^V
\end{aligned}
\end{eqnarray}
where ${\M W}^Q, {\M W}^K, {\M W}^V$, and ${\M W}^H \in \mathbb{R}^{d_t \times d_t}$ are the respective projection matrix for key, query, value, and similarity triangle encoding.

Since we conduct the attention in the candidate overlap region, ${\hat{p}}_i$ and ${\hat{q}}_i$ are candidate correspondences. The activated anchor point 'y' in ${\M g}_{yij}^{\hat{{P}}}$ and ${{\M g}_{yij}^{\hat{{Q}}}}$ may not be the corresponding anchor point. 'y' index denotes the third different anchor point selected by the max pooling from the source and target point cloud, respectively. This behavior is because the candidate overlap region may contain outliers. Then, we encode a new superpoint feature in the overlap region through the three interleaved attention layers above. 

\subsection{Feature Fusion and Propagation.}
Given the extraction of salient geometric features from the high-order transformer's output, we combine these features with the patch-level KPConv superpoint features using a linear layer. The resulting combination serves as input for the upsampling layers. We anticipate that our learned global scale high-order consistent geometric features will benefit the dense point matching step in the patch correspondence neighborhoods.

\begin{figure*}
	\centering
    \captionsetup[subfloat]{labelfont=scriptsize,textfont=scriptsize}
 \begin{minipage}{\textwidth}
        \centering
        \includegraphics[width=0.18\textwidth,height=2.3cm]{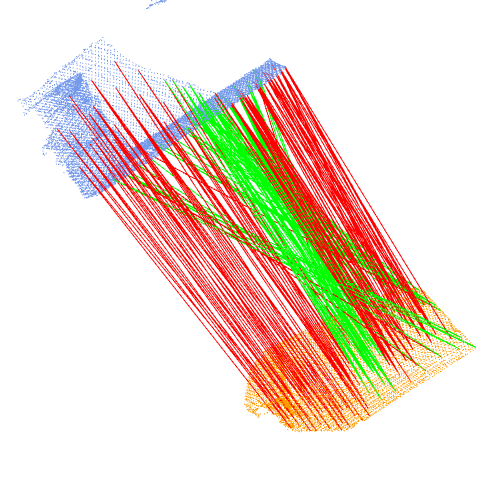}
        \includegraphics[width=0.18\textwidth,height=2.3cm]{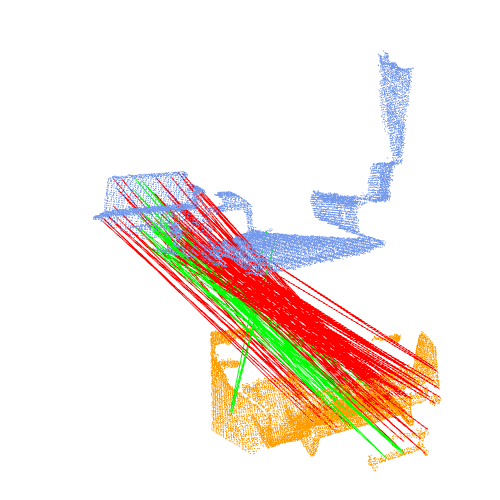}
        \includegraphics[width=0.18\textwidth,height=2.3cm]{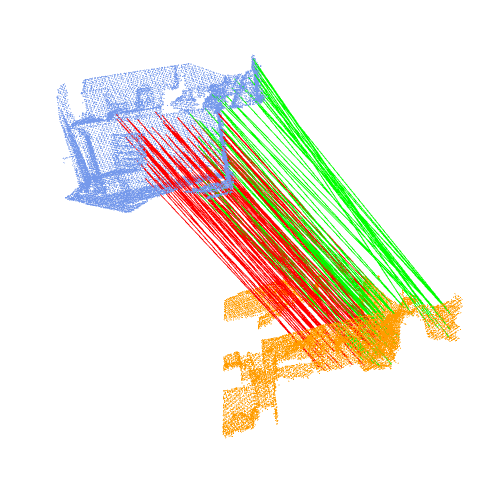}
        \includegraphics[width=0.18\textwidth,height=2.3cm]{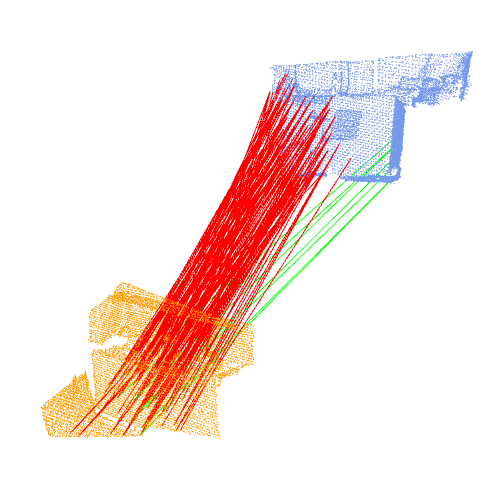}   
        \includegraphics[width=0.18\textwidth,height=2.3cm]{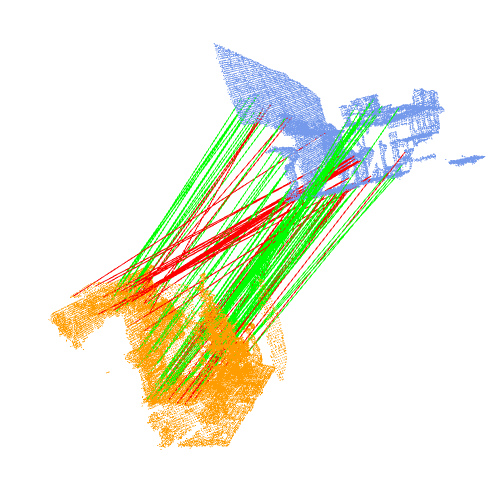}          
    \end{minipage}
    
    
    \begin{minipage}{\textwidth}
        \centering
        \includegraphics[width=0.18\textwidth,height=2.3cm]{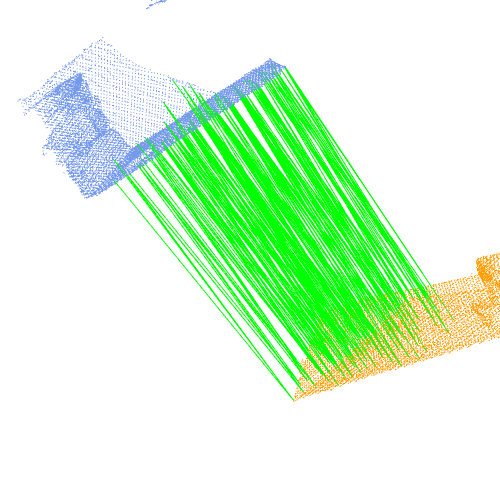}
        \includegraphics[width=0.18\textwidth,height=2.3cm]{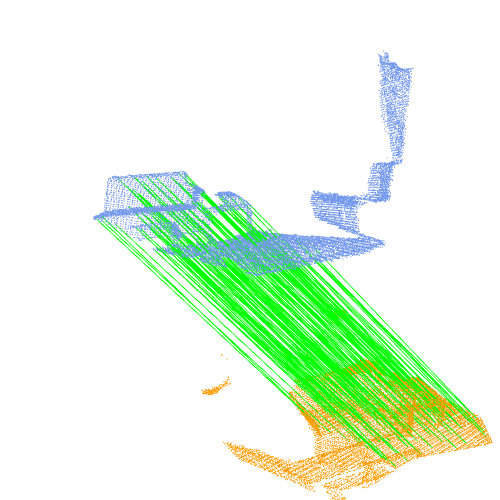}
        \includegraphics[width=0.18\textwidth,height=2.3cm]{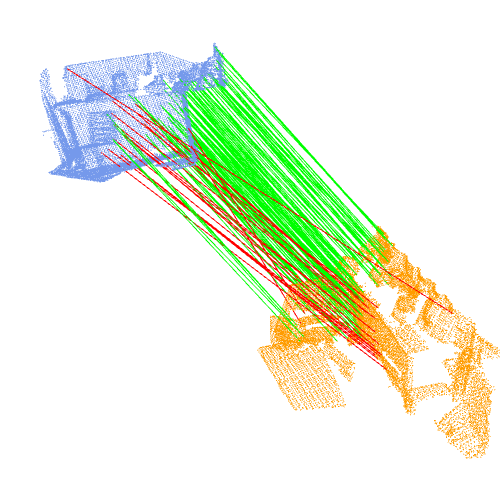}
        \includegraphics[width=0.18\textwidth,height=2.3cm]{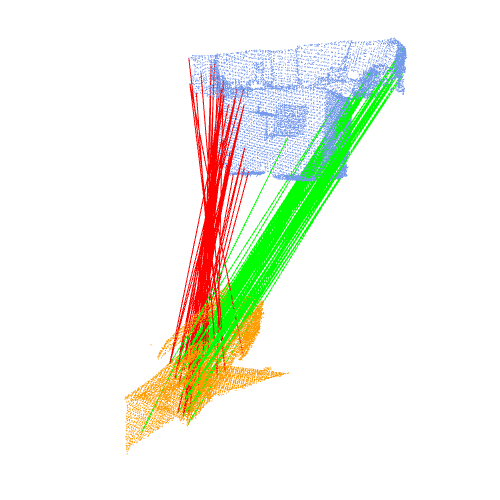}
        \includegraphics[width=0.18\textwidth,height=2.3cm]{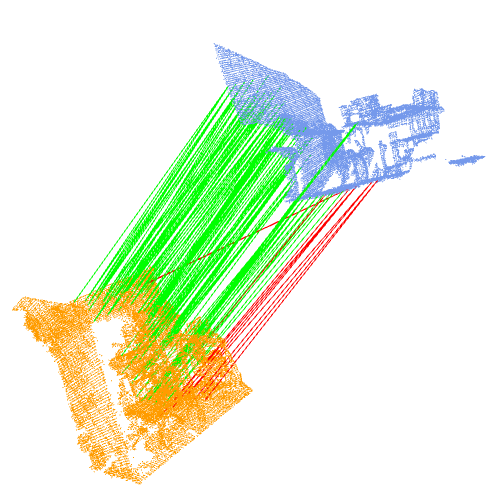}        
    \end{minipage}

	\caption{The visualizations showcase the patch correspondences achieved by our model on the 3DMatch dataset. The registration results in the top line were generated using GeoTR\cite{qin2022geometric}+LGR, while the results in the bottom line were generated using GeoTR~\cite{qin2022geometric} + High-Order Transformer + LGR. The \textcolor{green}{inlier}/\textcolor{red}{outlier} correspondences are highlighted in \textcolor{green}{green}/\textcolor{red}{red}. These visualizations serve to demonstrate the effectiveness of our high-order transformer in eliminating geometric inconsistencies in matches. Please zoom in for more detailed observations.}
	\label{res:vis_ho}
\end{figure*}

\begin{figure}
	\centering
    \captionsetup[subfloat]{labelfont=scriptsize,textfont=scriptsize}
    \subfloat[GeoTR~\cite{qin2022geometric} + LGR ]{\includegraphics[width=0.4\linewidth,height=2.3cm]{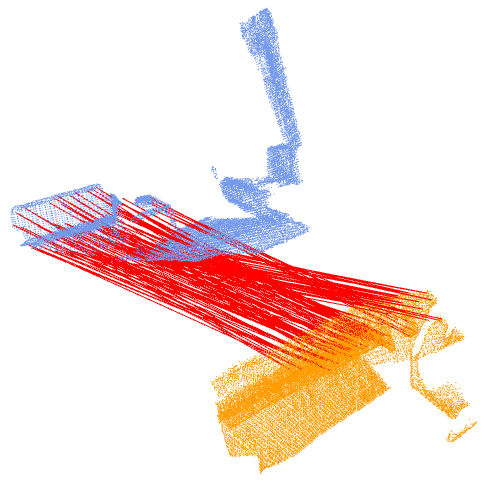}}\quad
	\subfloat[GeoTR~\cite{qin2022geometric} + HO + LGR ]{\includegraphics[width=0.4\linewidth,height=2.3cm]{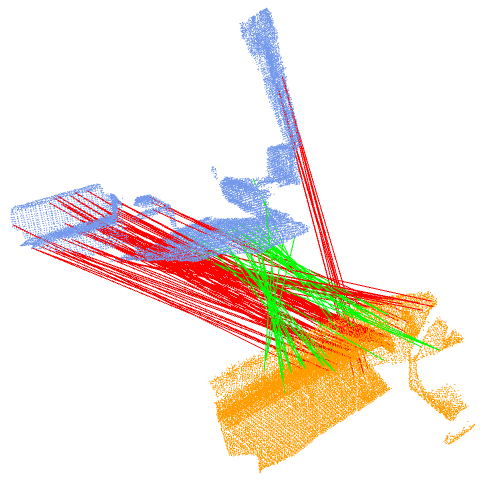}}\quad
    \subfloat[GeoTR~\cite{qin2022geometric} + HO + AUG + LGR]{\includegraphics[width=0.4\linewidth,height=2.3cm]{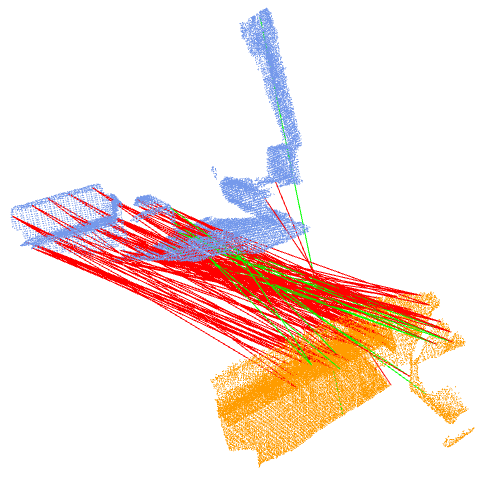}}\quad
    \subfloat[GeoTR~\cite{qin2022geometric} + HO + AUG + SE + LGR]{\includegraphics[width=0.35\linewidth,trim=30 90 10 10,clip]{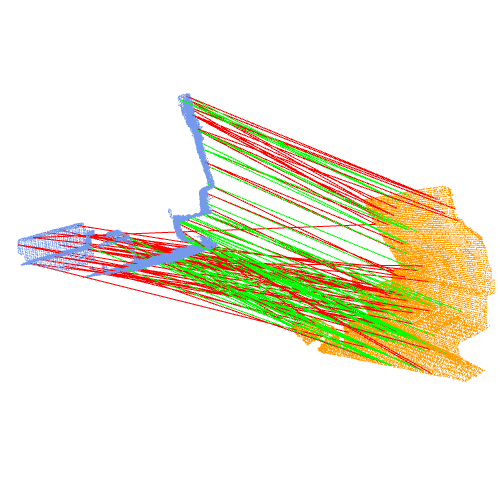}}\quad
   
    \caption{This visualization shows that the high-level semantic encoder captures the semantic information of the wall. HO, AUG, and SE denote the high-order transformer, superpoint augmentation, and high-level semantic encoder. The \textcolor{green}{inlier}/\textcolor{red}{outlier} correspondences are highlighted in \textcolor{green}{green}/\textcolor{red}{red}. Please zoom in for more detailed observations.}
    \label{res:ho_aug_sem}
\end{figure}

\section{Experiments}
\subsection{Implementation Details}
We utilize the overlap-aware circle loss~\cite{qin2022geometric} and point matching loss~\cite{qin2022geometric} to supervise the super point and dense point feature learning. Following~\cite{qin2022geometric}, we evaluate our proposed model on the indoor dataset 3DMatch~\cite{zeng20173dmatch} and outdoor dataset KITTI Odometry Benchmark~\cite{geiger2013vision}.

The proposed model is trained and tested with PyTorch~\cite{paszke2019pytorch} on one NVIDIA RTX 3090 GPU. We train our model with Adam optimizer\cite{loshchilov2017decoupled} with 30 epochs on 3DMatch\cite{zeng20173dmatch} and 150 epochs on KITTI\cite{geiger2013vision}. The learning rate, learning decay, learning rate decay steps, and weight decay are set to 1e-4/1e-4, 0.95/0.95, 1/4, and 1e-6/1e-6 for 3DMatch/KITTI, respectively. 3DMatch and KITTI's matching radii are set to 5cm and 30cm, respectively. The number of super point correspondences $N_c$ is set to 256 for all experiments. The number of farthest anchor points K is set to 64. The number of attention layers in high order geometric transformer ($N_g$) and semantic encoder ($N_s$) is 6 and 3, respectively. The threshold values $\lambda_{10}$ and $\lambda_{21}$ in salient super points augmentation are set to 0.6. The dimension of super point features $d_t$ is set to 1024 for 3DMatch/KITTI. $r_{salient}$ used in \ref{approach}.\textit{D} is set to 0.15/3.0 for 3DMatch/KITTI.

\subsection{Indoor Dataset: 3DMatch\&3DLoMatch}

\subsubsection{Dataset}

3DMatch~\cite{zeng20173dmatch} dataset has 62 scenes. They are split into 46 for training, 8 for validation, and 8 for testing. The overlap ratio of the scanned pairs in 3DMatch~\cite{zeng20173dmatch} and 3DLoMatch~\cite{huang2021predator} are higher and lower than 0.3, respectively.

\subsubsection{Evaluation Protocol}
Following~\cite{huang2021predator,qin2022geometric}, we utilize the following protocols to evaluate our model. (1) Inlier Ration(IR): The fraction of correct correspondences whose distance is under a threshold (i.e., 0.1m) under the ground
truth transformation. (2) Feature Matching Recall (FMR): the ratio of scanned pairs with the inlier ratio above a threshold (i.e., 5$\%$). (3) Registration Recall (RR): The fraction of successful registered point cloud pairs with a predicted transformation error under a threshold (i.e., RMSE$<$0.2). (4) Relative Rotation Error~(RRE) and Relative Translation
Error~(RTE): the error between the predicted transformation and the ground truth.
\subsubsection{Results}
We compare our model with feature matching methods:  FCGF~\cite{choy2019fully}, D3Feat~\cite{bai2020d3feat}, Predator~\cite{huang2021predator}, CoFiNet~\cite{yu2021cofinet}, RIGA~\cite{yu2022riga}, RegTR\cite{yew2022regtr}, Lepard~\cite{li2022lepard}, GeoTR~\cite{qin2022geometric}, and RoITr~\cite{yu2023rotation}. We utilize the local-to-global (LGR~\cite{bai2021pointdsc}) estimator for GeoTR~\cite{qin2022geometric} and our model since LGR is a more deterministic, stable, and faster method than the RANSAC tool. The results in TAB.\ref{res:3dmatch} show that our model achieves the best registration recall on 3DMatch/3DLoMatch with $93.8\%$/$76.5\%$. While our model prioritizes identifying salient and meaningful points necessary for successful registration, it may dilute meaningless and unnecessarily clustered inlier points, leading to lower scores in the IR metric.

\subsubsection{Ablation Studies}
We utilize the GeoTR~\cite{qin2022geometric} as our backbone in the ablation study experiments. We augment the super points to use the local salient points as super points. Then, we employ a new high-level semantic encoder to encode object-level semantics as inputs of local and low-oder geometric transformers~\cite{qin2022geometric}. After that, in the candidate overlap region, we employ our newly designed high-order geometric transformer to encode the inter-frame salient geometric feature and propagate it up to dense points to improve the feature learning of the backbone. As shown in TAB.\ref{res:ablay}, the high-order geometric transformer improves the registration recall on 3DMatch/3DLoMatch by about 0.8/0.6 percent points (pp). By exploiting the salient super point, augmentation increases the registration recall by about 0.6/1.0 pp on 3DMatch/3DLoMatch. After employing the high-level semantic encoder, our model can improve the registration recall by about 1.2/0.9 pp on 3DMatch/3DLoMatch.

\textbf{High-order Geometric Transformer.} We provide visualizations to demonstrate the effectiveness of our model. The super point matching matrices in GeoTR\cite{qin2022geometric} are computed by the inner product of point features. However, since the self-attention of GeoTR~\cite{qin2022geometric} operates on both overlap region and non-overlap region, it has several ambiguities matches in super point matching, as shown in the top line of Fig.\ref{res:vis_ho}. However, in our model, the super points learn the high-order salient consistent geometric features. As shown in the bottom line of Fig.\ref{res:vis_ho}, by employing the high-order transformer in the candidate overlap region, the outlier correspondences are eliminated, and the more inlier correspondences are identified. 

\textbf{High-level Semantic Encoder.} In this section, we showcase the effectiveness of the semantic encoder. As shown in (d) of Fig.\ref{res:ho_aug_sem}, the semantic encoder effectively captures the semantic information of the small table to the left sofa and the partial scan of the sofa adjacent to the wall. The correspondences of the sofa (adjacent to the wall), small desk, and walls are captured. Due to the flatness of the location on the wall, the shape signature is not very distinctive, resulting in less pronounced positive effects of superpoints augmentation. In contrast, in (a) of Fig.\ref{res:ho_aug_sem}, the correspondences of the wall are entirely lost. The results demonstrate that the high-level semantic encoder is crucial in embedding object-level information into these super points. Together with the high-order geometric transformer, it significantly contributes to the outstanding performance achieved by the final version of our model.

\begin{figure}[t]
   \centering{\includegraphics[width=6cm, height=4.5cm]{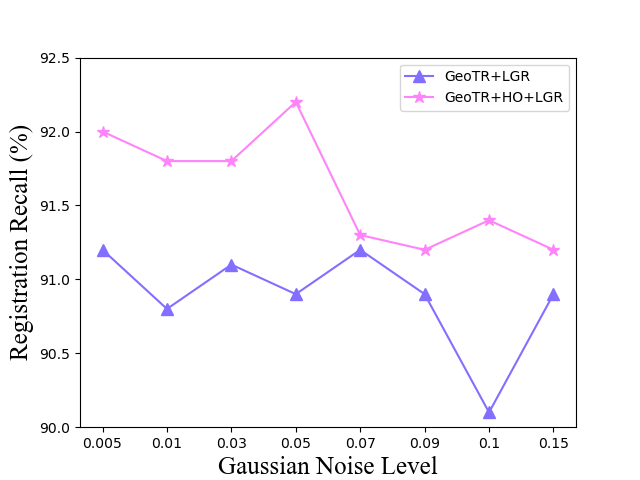}}
    \setlength{\abovecaptionskip}{-0.1cm} 
	\caption{The robustness of high-order geometric transformer against the Gaussian noise of point clouds inputs. Zoom in for details.}
    \label{noise_distillation}
\end{figure}

\subsubsection{Noise Distillation of High-order Geometric Transformer}
To further prove the high-order geometric transformer's robustness against the sensor noise, we experiment with different levels of Gaussian noise added to point clouds on the 3DMatch dataset. The results in Fig. \ref{noise_distillation} show that our high-order geometric transformer enhances the robustness of GeoTR\cite{qin2022geometric} at different sensor noise levels.

\subsubsection{Computational Complexity}
The running time of the super point augmentation module, high-order geometric transformer, and semantic encoder are listed in TAB.\ref{res:comlexsity}.

\begin{table}
\caption{
Quantitative results on the 3DMatch and 3DLoMatch benchmarks. The best results are highlighted in bold, and the second-best results are underlined. In this paper, we focus on the improvement over the previous SoTA method GeoTR\cite{qin2022geometric}. $*$ denotes that we only give a comparison on the original 3DMatch dataset.}

\label{res:3dmatch}

\footnotesize
    \setlength{\tabcolsep}{1.0pt}
 
\begin{tabular}{c|ccccc}

            \toprule
             \midrule
         
\multirow{2}{*}{Method}  & &\multicolumn{2}{c}{3DMatch}  \\
                  &       RR (\%)$\uparrow$             &         IR (\%)$\uparrow$            &           FMR (\%)$\uparrow$        &RRE ($^\circ$)$\downarrow$&RTE (m)$\downarrow$ \\
                  \midrule
                  FCGF\cite{choy2019fully}&83.3&48.7 &97.0&1.949&0.066\\
                  D3Feat~\cite{bai2020d3feat}&83.4&40.4 &94.5&2.161&0.067\\
                  Predator~\cite{huang2021predator}&90.6&57.1&96.5&2.029&0.064\\
                  CoFiNet~\cite{yu2021cofinet}&88.4&51.9 &\uline{98.1}&2.011&0.067\\
                  RIGA\cite{yu2022riga}&89.3&68.4&97.9&-&-\\
                  RegTR\cite{yew2022regtr}&92.0&-&-&\textbf{1.570}&\textbf{0.049}\\
                  Lepard\cite{li2022lepard}&\uline{93.5}&-&-&2.480 &0.072\\  

                  RoITr$^{\ast}$\cite{yu2023rotation}&91.9&\textbf{82.6}&98.0&-&-\\
                  \rowcolor[gray]{0.9} 
                  GeoTR+LGR\cite{qin2022geometric}&91.2&\uline{70.1} &\uline{98.1}&\uline{1.745}&\uline{0.061}\\   
                  \rowcolor[gray]{0.9} 
                  SGNet+LGR (ours) &\textbf{93.8}&41.4& \textbf{98.4}&1.752&\uline{0.061}\\  
                  \midrule
                 \multirow{2}{*}{Method}  & &\multicolumn{2}{c}{3DLoMatch}  \\
                 &       RR (\%)$\uparrow$             &         IR (\%)$\uparrow$            &           FMR (\%)$\uparrow$        &RRE ($^\circ$)$\downarrow$&RTE (m)$\downarrow$ \\
                  \midrule
                  FCGF\cite{choy2019fully}&38.2&17.2 &74.2&3.147&0.100\\
                  D3Feat~\cite{bai2020d3feat}&46.9&14.0 &67.0&3.361&0.103\\
                  Predator~\cite{huang2021predator}&61.2&28.3&76.3&3.048&0.093\\
                  CoFiNet~\cite{yu2021cofinet}&64.2&26.7 &83.3&3.280&0.094\\
                  RIGA\cite{yu2022riga}&65.1&32.1&85.1&-&-\\
                  RegTR\cite{yew2022regtr}&64.8&-&-&\uline{2.830}&0.077\\
                  Lepard\cite{li2022lepard}&69.0&-&-&4.100 &0.108\\        
                  RoITr$^{\ast}$\cite{yu2023rotation}&\uline{74.8}&\textbf{54.3}&\textbf{89.6}&-&-\\
                  \rowcolor[gray]{0.9}        
                  GeoTR+LGR\cite{qin2022geometric}&74.0&\uline{41.0} &\uline{87.7}&2.886&\uline{0.088}\\    
                  \rowcolor[gray]{0.9} 
                  SGNet+LGR (ours)&\textbf{76.5}&19.2&83.4 &\textbf{2.767}&\textbf{0.085}\\ 
		  \bottomrule
\end{tabular}
\end{table}

\begin{table}
	\caption{Ablation of the network architecture on 3DMatch/3DLoMatch benchmark. AUG, HOTR, and HSE denote the super point augmentation, high-order geometric transformer, and high-level semantic encoder}.
	\label{res:ablay}
	\centering
	\resizebox{0.49\textwidth}{!}{
    \begin{tabular}{ccc|ccc|ccc}
		\toprule
  \toprule
   
        \multicolumn{2}{l}{} &  &3DMatch &    &  &3DLoMatch&  \\
        \midrule
        AUG&HOTR&HSE&RR (\%)$\uparrow$&RRE ($^\circ$)$\downarrow$&RTE (cm)$\downarrow$&RR (\%)$\uparrow$&RRE ($^\circ$)$\downarrow$&RTE (m)$\downarrow$\\
        \midrule 
          &&&91.2&1.745&0.061&74.0&2.886&0.088\\
          &          $\surd$&& 92.0 & 1.791&0.062& 74.6&2.841&0.088\\
          &          $\surd$&$\surd$& 92.2 & 2.178&0.073&75.1&2.825&0.087\\          $\surd$&$\surd$&&92.6&1.774&0.062&75.6&2.803&0.085 \\
          $\surd$&$\surd$&$\surd$&93.8&1.752&0.061&76.5&2.767&0.085 \\
        \bottomrule
    \end{tabular}
    }
\end{table}

\begin{table}
	\caption{Runtime of each component averaged over 1623 fragment pairs of 3DMatch in milli-seconds.}
	\label{res:comlexsity}
	\centering
	 \resizebox{0.99\linewidth}{!}{
	\begin{tabular}{c|ccccccc}
		\toprule
   \midrule
		Method&data loader&encoder&attention layer&decoder&AUG&HOTR&HSE\\
		\midrule
        Predator\cite{huang2021predator}&191& 9& 70&1& \XSolidBrush&\XSolidBrush&\XSolidBrush\\
        GeoTR\cite{qin2022geometric}&-& -& 60&1& \XSolidBrush&\XSolidBrush&\XSolidBrush\\
        SGNet+LGR&-&-&60&-&10&50&30\\
		\bottomrule
	\end{tabular}}
\end{table}

\begin{table}
	\caption{Quantitative results on the KITTI odometry benchmark. The best results are highlighted in bold, and the second-best results are underlined.}
	\label{res:KITTI}
	\centering
	\resizebox{0.99\linewidth}{!}{
	\begin{tabular}{c|ccc}
		\toprule
	 \midrule
		Method&RTE (cm)$\downarrow$&RRE ($^\circ$)$\downarrow$&RR ($\%$)$\uparrow$\\
		\midrule
		FMR\cite{huang2020feature}&66&1.49&90.6\\
        DGR\cite{choy2020deep}&32&0.37&98.7\\
        HRegNet\cite{lu2021hregnet}&12&0.29&\uline{99.7}\\  
		3DFeat-Net\cite{yew20183dfeat} &25.9 &\uline{0.25}& 96.0\\
		FCGF\cite{choy2019fully}&9.5&0.30&96.6\\
		D3Feat\cite{bai2020d3feat}&7.2&0.30&\textbf{99.8}\\
		SpinNet\cite{ao2021spinnet}&9.9&0.47&99.1\\
        Predator\cite{huang2021predator}&\uline{6.8}&0.27&\textbf{99.8}\\
		CoFiNet\cite{yu2021cofinet}&8.2&0.41&\textbf{99.8}\\
		GeoTR+LGR\cite{qin2022geometric}&7.4&0.27& \textbf{99.8}\\
		SGNet+LGR (Ours)&\textbf{5.4}&\textbf{0.24}&\textbf{99.8}\\
		\bottomrule
	\end{tabular}}
\end{table}

\subsection{Outdoor Dataset: KITTI Odometry}
\subsubsection{Dataset} KITTI Odometry benchmark~\cite{geiger2013vision} consists of an 11-point cloud sequence scanned by a LiDAR sensor. Following~\cite{bai2020d3feat,huang2021predator,choy2019fully}, we split the sequences to 0-5/6-7/8-10 for training/validation/testing. In line with~\cite{bai2020d3feat,choy2019fully,huang2021predator,qin2022geometric}, we refine the ground truth transformation by ICP~\cite{besl1992method} and only use point cloud pairs that are
at most 10m away from each other for evaluation.

\subsubsection{Evaluation Protocol}
As in~\cite{qin2022geometric}, we utilize three metrics to evaluate our model: (1) Registration Recall~(RR), the fraction of successful registration pairs ($i.e.,$ RRE$<5^\circ$ and RTE$<$2m). The other two metrics, Relative Rotation Error~(RRE) and Relative Translation Error~(RTE) go with the same used in the 3DMatch benchmark.

\subsubsection{Results}
We compare our model with several state-of-the-art methods: FMR\cite{huang2020feature}, DGR\cite{choy2020deep}, HRegNet\cite{lu2021hregnet}, FCGF\cite{choy2019fully},  D3Feat\cite{bai2020d3feat},  SpinNet\cite{ao2021spinnet}, Predator\cite{huang2021predator},  CoFiNet\cite{yu2021cofinet}, and GeoTR\cite{qin2022geometric}. As shown in TAB.\ref{res:KITTI}, the quantitative results show that our method can handle outdoor scene registration and achieve competitive performance.

\subsection{Limitations.}
As shown in TAB.\ref{res:3dmatch}, SGNet's IR metric is relatively lower compared to other baselines, primarily due to the inlier clustering phenomenon\cite{huang2021predator}. However, because our method focuses on key points that enhance the recall metric, the clustered inlier points become less critical. SGNet emphasizes high-order geometric consistency features to ensure registration success, with the high-order saliency transformer selecting points from the candidate correspondences generated by the semantic-aware geometric encoder. Consequently, the repeatability and precision metrics\cite{li2019usip,yew20183dfeat} of SGNet may be lower compared to methods that rely on semantic keypoints\cite{li2019usip,yew20183dfeat}.

\section{Conclusion}
This paper proposes a novel approach for point cloud registration that overcomes limitations in existing methods. By incorporating object-level semantic information through a high-level Semantic Encoder (SE) and leveraging intrinsic shape signatures to select meaningful super points, our approach improves the accuracy of super point matching. It enhances the discriminative nature of the selected points. We have also introduced a High-Order (HO) geometric consistency transformer that identifies salient points within candidate overlap regions and a unique anchor node selection strategy to improve its performance. Experimental results on outdoor and indoor datasets demonstrate the effectiveness of our approach in achieving accurate and competitive point cloud registration results.

{\normalem
\bibliographystyle{IEEEtran} 
\bibliography{root}

\begin{thebibliography}{10}
\providecommand{\url}[1]{#1}
\csname url@samestyle\endcsname
\providecommand{\newblock}{\relax}
\providecommand{\bibinfo}[2]{#2}
\providecommand{\BIBentrySTDinterwordspacing}{\spaceskip=0pt\relax}
\providecommand{\BIBentryALTinterwordstretchfactor}{4}
\providecommand{\BIBentryALTinterwordspacing}{\spaceskip=\fontdimen2\font plus
\BIBentryALTinterwordstretchfactor\fontdimen3\font minus \fontdimen4\font\relax}
\providecommand{\BIBforeignlanguage}[2]{{%
\expandafter\ifx\csname l@#1\endcsname\relax
\typeout{** WARNING: IEEEtran.bst: No hyphenation pattern has been}%
\typeout{** loaded for the language `#1'. Using the pattern for}%
\typeout{** the default language instead.}%
\else
\language=\csname l@#1\endcsname
\fi
#2}}
\providecommand{\BIBdecl}{\relax}
\BIBdecl

\bibitem{qin2022geometric}
Z.~Qin, H.~Yu, C.~Wang, Y.~Guo, Y.~Peng, and K.~Xu, ``Geometric transformer for fast and robust point cloud registration,'' in \emph{Proceedings of the IEEE/CVF Conference on Computer Vision and Pattern Recognition}, 2022, pp. 11\,143--11\,152.

\bibitem{huang2021comprehensive}
X.~Huang, G.~Mei, J.~Zhang, and R.~Abbas, ``A comprehensive survey on point cloud registration,'' 2021.

\bibitem{cadena2016past}
C.~Cadena, L.~Carlone, H.~Carrillo, Y.~Latif, D.~Scaramuzza, J.~Neira, I.~Reid, and J.~J. Leonard, ``Past, present, and future of simultaneous localization and mapping: Toward the robust-perception age,'' \emph{IEEE Transactions on robotics}, vol.~32, no.~6, pp. 1309--1332, 2016.

\bibitem{schonberger2016structure}
J.~L. Schonberger and J.-M. Frahm, ``Structure-from-motion revisited,'' in \emph{Proceedings of the IEEE conference on computer vision and pattern recognition}, 2016, pp. 4104--4113.

\bibitem{choi2015robust}
S.~Choi, Q.-Y. Zhou, and V.~Koltun, ``Robust reconstruction of indoor scenes,'' in \emph{Proceedings of the IEEE Conference on Computer Vision and Pattern Recognition}, 2015, pp. 5556--5565.

\bibitem{zhang2015visual}
J.~Zhang and S.~Singh, ``Visual-lidar odometry and mapping: Low-drift, robust, and fast,'' in \emph{2015 IEEE International Conference on Robotics and Automation (ICRA)}.\hskip 1em plus 0.5em minus 0.4em\relax IEEE, 2015, pp. 2174--2181.

\bibitem{thomas2019kpconv}
H.~Thomas, C.~R. Qi, J.-E. Deschaud, B.~Marcotegui, F.~Goulette, and L.~J. Guibas, ``Kpconv: Flexible and deformable convolution for point clouds,'' in \emph{Proceedings of the IEEE/CVF international conference on computer vision}, 2019, pp. 6411--6420.

\bibitem{bai2020d3feat}
X.~Bai, Z.~Luo, L.~Zhou, H.~Fu, L.~Quan, and C.-L. Tai, ``D3feat: Joint learning of dense detection and description of 3d local features,'' in \emph{Proceedings of the IEEE/CVF conference on computer vision and pattern recognition}, 2020, pp. 6359--6367.

\bibitem{huang2021predator}
S.~Huang, Z.~Gojcic, M.~Usvyatsov, A.~Wieser, and K.~Schindler, ``Predator: Registration of 3d point clouds with low overlap,'' in \emph{Proceedings of the IEEE/CVF Conference on computer vision and pattern recognition}, 2021, pp. 4267--4276.

\bibitem{yu2021cofinet}
H.~Yu, F.~Li, M.~Saleh, B.~Busam, and S.~Ilic, ``Cofinet: Reliable coarse-to-fine correspondences for robust pointcloud registration,'' \emph{Advances in Neural Information Processing Systems}, vol.~34, pp. 23\,872--23\,884, 2021.

\bibitem{vaswani2017attention}
A.~Vaswani, N.~Shazeer, N.~Parmar, J.~Uszkoreit, L.~Jones, A.~N. Gomez, {\L}.~Kaiser, and I.~Polosukhin, ``Attention is all you need,'' \emph{Advances in neural information processing systems}, vol.~30, 2017.

\bibitem{bai2021pointdsc}
X.~Bai, Z.~Luo, L.~Zhou, H.~Chen, L.~Li, Z.~Hu, H.~Fu, and C.-L. Tai, ``Pointdsc: Robust point cloud registration using deep spatial consistency,'' in \emph{Proceedings of the IEEE/CVF Conference on Computer Vision and Pattern Recognition}, 2021, pp. 15\,859--15\,869.

\bibitem{wang2021neural}
R.~Wang, J.~Yan, and X.~Yang, ``Neural graph matching network: Learning lawler’s quadratic assignment problem with extension to hypergraph and multiple-graph matching,'' \emph{IEEE Transactions on Pattern Analysis and Machine Intelligence}, 2021.

\bibitem{lee2011hyper}
J.~Lee, M.~Cho, and K.~M. Lee, ``Hyper-graph matching via reweighted random walks,'' in \emph{CVPR 2011}.\hskip 1em plus 0.5em minus 0.4em\relax IEEE, 2011, pp. 1633--1640.

\bibitem{zhong2009intrinsic}
Y.~Zhong, ``Intrinsic shape signatures: A shape descriptor for 3d object recognition,'' in \emph{2009 IEEE 12th international conference on computer vision workshops, ICCV workshops}.\hskip 1em plus 0.5em minus 0.4em\relax IEEE, 2009, pp. 689--696.

\bibitem{yu2023rotation}
H.~Yu, Z.~Qin, J.~Hou, M.~Saleh, D.~Li, B.~Busam, and S.~Ilic, ``Rotation-invariant transformer for point cloud matching,'' in \emph{Proceedings of the IEEE/CVF Conference on Computer Vision and Pattern Recognition}, 2023, pp. 5384--5393.

\bibitem{deng2018ppf}
H.~Deng, T.~Birdal, and S.~Ilic, ``Ppf-foldnet: Unsupervised learning of rotation invariant 3d local descriptors,'' in \emph{Proceedings of the European Conference on Computer Vision (ECCV)}, 2018, pp. 602--618.

\bibitem{chen2022sc2}
Z.~Chen, K.~Sun, F.~Yang, and W.~Tao, ``Sc2-pcr: A second order spatial compatibility for efficient and robust point cloud registration,'' in \emph{Proceedings of the IEEE/CVF Conference on Computer Vision and Pattern Recognition}, 2022, pp. 13\,221--13\,231.

\bibitem{zhang20233d}
X.~Zhang, J.~Yang, S.~Zhang, and Y.~Zhang, ``3d registration with maximal cliques,'' in \emph{Proceedings of the IEEE/CVF Conference on Computer Vision and Pattern Recognition}, 2023, pp. 17\,745--17\,754.

\bibitem{yu2023peal}
J.~Yu, L.~Ren, Y.~Zhang, W.~Zhou, L.~Lin, and G.~Dai, ``Peal: Prior-embedded explicit attention learning for low-overlap point cloud registration,'' in \emph{Proceedings of the IEEE/CVF Conference on Computer Vision and Pattern Recognition}, 2023, pp. 17\,702--17\,711.

\bibitem{li2019usip}
J.~Li and G.~H. Lee, ``Usip: Unsupervised stable interest point detection from 3d point clouds,'' in \emph{Proceedings of the IEEE/CVF international conference on computer vision}, 2019, pp. 361--370.

\bibitem{harris1988combined}
C.~Harris, M.~Stephens \emph{et~al.}, ``A combined corner and edge detector,'' in \emph{Alvey vision conference}, vol.~15, no.~50.\hskip 1em plus 0.5em minus 0.4em\relax Citeseer, 1988, pp. 10--5244.

\bibitem{lowe2004distinctive}
D.~G. Lowe, ``Distinctive image features from scale-invariant keypoints,'' \emph{International journal of computer vision}, vol.~60, pp. 91--110, 2004.

\bibitem{zeng20173dmatch}
A.~Zeng, S.~Song, M.~Nie{\ss}ner, M.~Fisher, J.~Xiao, and T.~Funkhouser, ``3dmatch: Learning local geometric descriptors from rgb-d reconstructions,'' in \emph{Proceedings of the IEEE conference on computer vision and pattern recognition}, 2017, pp. 1802--1811.

\bibitem{geiger2013vision}
A.~Geiger, P.~Lenz, C.~Stiller, and R.~Urtasun, ``Vision meets robotics: The kitti dataset,'' \emph{The International Journal of Robotics Research}, vol.~32, no.~11, pp. 1231--1237, 2013.

\bibitem{paszke2019pytorch}
A.~Paszke, S.~Gross, F.~Massa, A.~Lerer, J.~Bradbury, G.~Chanan, T.~Killeen, Z.~Lin, N.~Gimelshein, L.~Antiga \emph{et~al.}, ``Pytorch: An imperative style, high-performance deep learning library,'' \emph{Advances in neural information processing systems}, vol.~32, 2019.

\bibitem{loshchilov2017decoupled}
I.~Loshchilov and F.~Hutter, ``Decoupled weight decay regularization,'' \emph{arXiv preprint arXiv:1711.05101}, 2017.

\bibitem{choy2019fully}
C.~Choy, J.~Park, and V.~Koltun, ``Fully convolutional geometric features,'' in \emph{Proceedings of the IEEE/CVF International Conference on Computer Vision}, 2019, pp. 8958--8966.

\bibitem{yu2022riga}
H.~Yu, J.~Hou, Z.~Qin, M.~Saleh, I.~Shugurov, K.~Wang, B.~Busam, and S.~Ilic, ``Riga: Rotation-invariant and globally-aware descriptors for point cloud registration,'' \emph{arXiv preprint arXiv:2209.13252}, 2022.

\bibitem{yew2022regtr}
Z.~J. Yew and G.~H. Lee, ``Regtr: End-to-end point cloud correspondences with transformers,'' in \emph{Proceedings of the IEEE/CVF Conference on Computer Vision and Pattern Recognition}, 2022, pp. 6677--6686.

\bibitem{li2022lepard}
Y.~Li and T.~Harada, ``Lepard: Learning partial point cloud matching in rigid and deformable scenes,'' in \emph{Proceedings of the IEEE/CVF Conference on Computer Vision and Pattern Recognition}, 2022, pp. 5554--5564.

\bibitem{huang2020feature}
X.~Huang, G.~Mei, and J.~Zhang, ``Feature-metric registration: A fast semi-supervised approach for robust point cloud registration without correspondences,'' in \emph{Proceedings of the IEEE/CVF conference on computer vision and pattern recognition}, 2020, pp. 11\,366--11\,374.

\bibitem{choy2020deep}
C.~Choy, W.~Dong, and V.~Koltun, ``Deep global registration,'' in \emph{Proceedings of the IEEE/CVF conference on computer vision and pattern recognition}, 2020, pp. 2514--2523.

\bibitem{lu2021hregnet}
F.~Lu, G.~Chen, Y.~Liu, L.~Zhang, S.~Qu, S.~Liu, and R.~Gu, ``Hregnet: A hierarchical network for large-scale outdoor lidar point cloud registration,'' in \emph{Proceedings of the IEEE/CVF International Conference on Computer Vision}, 2021, pp. 16\,014--16\,023.

\bibitem{yew20183dfeat}
Z.~J. Yew and G.~H. Lee, ``3dfeat-net: Weakly supervised local 3d features for point cloud registration,'' in \emph{Proceedings of the European conference on computer vision (ECCV)}, 2018, pp. 607--623.

\bibitem{ao2021spinnet}
S.~Ao, Q.~Hu, B.~Yang, A.~Markham, and Y.~Guo, ``Spinnet: Learning a general surface descriptor for 3d point cloud registration,'' in \emph{Proceedings of the IEEE/CVF conference on computer vision and pattern recognition}, 2021, pp. 11\,753--11\,762.

\bibitem{besl1992method}
P.~J. Besl and N.~D. McKay, ``Method for registration of 3-d shapes,'' in \emph{Sensor fusion IV: control paradigms and data structures}, vol. 1611.\hskip 1em plus 0.5em minus 0.4em\relax Spie, 1992, pp. 586--606.

\end{thebibliography}
}

\newpage

\vfill

\end{document}